\documentclass[runningheads]{llncs}
\usepackage[T1]{fontenc}
\usepackage{graphicx}
\usepackage{algpseudocode}
\usepackage{algorithm}
\usepackage{array}
\usepackage{hyperref}
\usepackage{multirow}
\usepackage{booktabs}
\usepackage{amsmath}

\usepackage[capitalize]{cleveref}
\usepackage{amssymb}

\usepackage[LGR,T1]{fontenc}

\begin{document}
\title{The Influence of Iconicity in Transfer Learning for Sign Language Recognition}
\author{Keren Artiaga\inst{1}\thanks{\noindent
This version of the contribution has been accepted for publication, after peer review but is not the Version of Record and does not reflect post-acceptance improvements, or any corrections. The Version of Record is available online at: \url{https://doi.org/10.1007/978-3-031-70239-6_16}. Use of this Accepted Version is subject to the publisher’s Accepted Manuscript terms of use: \url{https://www.springernature.com/gp/open-research/policies/accepted-manuscript-terms}.
} \and
Conor Lynch\inst{2} \and
Haithem Afli\inst{1} \and
Mohammed Hasanuzzaman\inst{1,3}
}
\authorrunning{K. Artiaga et al.}
\institute{
ADAPT Centre, Munster Technological University, Cork, Ireland \\
\email{keren.artiaga@adaptcentre.ie, haithem.afli@adaptcentre.ie, mohammed.hasanuzzaman@adaptcentre.ie} \and
Nimbus Centre, Munster Technological University, Cork, Ireland \\
\email{conor.lynch@mtu.ie} \and
School of Electronics, Electrical Engineering and Computer Science (EEECS), 
Queen's University Belfast, UK
}
\maketitle              % typeset the header of the contribution
\begin{abstract}
Most sign language recognition research relies on Transfer
Learning (TL) from vision-based datasets such as ImageNet. Some extend this to alternatively available language datasets, often focusing on
signs with cross-linguistic similarities. This body of work examines the
necessity of these likenesses on effective knowledge transfer by comparing
TL performance between iconic signs of two different sign language pairs:
Chinese to Arabic and Greek to Flemish. Google Mediapipe was utilised
as an input feature extractor, enabling spatial information of these signs
to be processed with a Multilayer Perceptron architecture and the temporal information with a Gated Recurrent Unit. Experimental results
showed a 7.02\% improvement for Arabic and 1.07\% for Flemish when
conducting iconic TL from Chinese and Greek respectively.

\end{abstract}
\section{Introduction}
Sign language datasets, often limited in size compared to spoken language datasets, are susceptible to overfitting \cite{tarres2023sign}, \cite{marivate2020investigating}. Sign Language Recognition (SLR) studies often work with datasets containing fewer than 30k samples \cite{SultanMakramKayedAli+2022+191+210}. This is a stark contrast to spoken language, where a language pair corpora with less than 500k samples is considered low-resourced \cite{lakew-etal-2019-adapting}, \cite{liu-etal-2020-multilingual-denoising}.
To address this issue, Transfer Learning (TL) has gained traction in SLR and Sign Language Translation (SLT) - typically involving knowledge transfer from ImageNet, a large vision-based dataset \cite{Das2022AHA}, \cite{10063044}, \cite{9115767}. Recent studies have explored TL between different language datasets, for instance from American Sign Language (ASL) to Ankara University Turkish Sign Language (AUTSL), the Spanish: Lengua de Signos Española (LSE) kinesics to AUTSL \cite{9523152} and British Sign Language (BSL) to ASL \cite{Bird2020BritishSL} - where the improved recognition is explained by their shared similarities. However, BSL and ASL belong to different language families and their similarities can be attributed to \emph{iconicity} \cite{visualiconicity} - a semiotic concept where a word or sign resembles its meaning, such as the sign or gesture for \emph{Think} which always involves a mime or hand movement towards the head across various languages. \cite{stokoe1976dictionary} demonstrated in their 1976 study that ASL signs possess 25\% pantomimic or iconic attributes. Furthermore, in a more recent 2018 study by \cite{10.3389/fpsyg.2018.01433}, a significant 68\% iconicity correlation between 604 BSL and 993 ASL signs was reported. In addition to the aforementioned, quantitative analyses have also confirmed cross-linguistic similarities in iconic signs across various sign languages \cite{visualiconicity}. This study builds upon these findings by conducting transfer learning between iconic signs, exploring the impact of movement differences on knowledge transfer in SLR - a topic, to the authors' knowledge, that has been not previously explored.

The novel contribution of this research work involves a comparative analysis of TL from iconic signs across datasets of two disxtinct sign language source/target pairs: 

(1) From SLR500\footnote{\href {https://ustc-slr.github.io/datasets/2015_csl/}{https://ustc-slr.github.io/datasets/2015\_csl}}, a Chinese Sign Language (CSL) SLR dataset comprising 500 daily Chinese sign categories to KArSL\footnote{\href {https://hamzah-luqman.github.io/KArSL}{https://hamzah-luqman.github.io/KArSL/}} (KFUPM Arabic Sign Language), an Arabic Sign Language (ArSL) database consisting of 502 sign words belonging to categories such as Letters, Numbers, Health, and Family among others. Each sign from SLR500 is performed 5 times by 50 signers, whereas for KArSL, three signers performed each sign 50 times.

(2) From an isolated Greek Sign Language (GSL)\footnote{\href{https://vcl.iti.gr/dataset/gsl/}{https://vcl.iti.gr/dataset/gsl/}}dataset that features cases of Deaf people interacting with police departments, hospitals, and citizen service centres to the Flemish Vlaamse Gebarentaal or VGT \footnote{\href{https://taalmaterialen.ivdnt.org/download/woordenboek-vgt/}{https://taalmaterialen.ivdnt.org/download/woordenboek-vgt/}} dataset. Woordenboek VGT is a collection of videos in Flemish Sign Language. 120 deaf people contributed to the Corpus VGT as informants. Age, region and gender were taken into account when selecting the participants. 

In particular, for GSL isolated, seven native signers performed the signs five times, whereas, for Woordenboek VGT, repetitions are present, but the authors did not indicate an exact number.

The experiments are designed so that the low-resource datasets are the target tasks while the higher-resource datasets are the source tasks. Therefore, CSL SLR500 and GSL isolated were chosen as the source languages due to their higher number of samples per class (250 and 54 respectively) relative to their corresponding lower-resource target corpus (150 and 13.64 respectively). Table \ref{tab: Overview of the iconic subsets} presents the number of classes for each subset along with the number of samples per class, as well as the iconic concepts these classes represent. These iconic concepts are identified in the study by \cite{visualiconicity} and are visualised in Figures \ref{fig:ostling_individualconcept} and \ref{fig:ostling_largeconcept} respectively. It is noteworthy that the first pair of subsets, CSL SLR500 and KArSL, share the same set of five iconic concepts (\emph{anatomy, hair, eyesight, love} and \emph{sound}), while the second pairing, GSL isolated and Woordenboek VGT, have merely three mutual iconic concepts -  \emph{anatomy, food} and \emph{sound}. This intended selection allows one to deduce whether the number of shared iconic concepts has a significant or otherwise impact on the transferability of knowledge.

\begin{table}[htbp!]
\centering
\resizebox{.75\textwidth}{!}{%
\centering

\begin{tabular}{llll}
\hline
\textbf{Dataset}                                                  & \textbf{\begin{tabular}[c]{@{}l@{}}Iconic \\ subset \\ no. of classes\end{tabular}} & \textbf{\begin{tabular}[c]{@{}l@{}}Iconic \\ subset\\ no. of samples\\ per class (mean)\end{tabular}} & \textbf{\begin{tabular}[c]{@{}l@{}}Iconic \\ concepts \\ the classes \\ belong to\end{tabular}}                            \\ \hline
CSL SLR500    \footnotemark[1]                                            & 8                                                                       & 250                                                                                        & \begin{tabular}[c]{@{}l@{}}Anatomy, Hair,\\ Eyesight, Love, Sound\end{tabular}                              \\
KArSL  \footnotemark[2]                                                   & 26                                                                      & 150                                                                                        & \begin{tabular}[c]{@{}l@{}}Anatomy, Hair, \\ Eyesight, Love, Sound\end{tabular}                             \\ \hline
GSL isolated   \footnotemark[3]                                           & 13                                                                      & 54                                                                                         & Anatomy, Food, Sound                                                                                        \\
\begin{tabular}[c]{@{}l@{}}Woordenboek\\ VGT \footnotemark[4] \end{tabular} & 85                                                                      & 13.64                                                                                      & \begin{tabular}[c]{@{}l@{}}Anatomy, Food, Sound,\\ Hair, Clothes, Eyesight,\\ Say, Love, Hear,\end{tabular} \\ \hline
\end{tabular}
}
\caption{Overview of the iconic subset}
\label{tab: Overview of the iconic subsets}
\vspace{-0.3cm}
{\footnotesize \footnotemark[1] \cite{zhou2021improving}, \cite{hu2023signbert+}, \cite{hu2021signbert}}
{\footnotesize \footnotemark[2] \cite{sidig2021karsl}}
{\footnotesize \footnotemark[3] \cite{GSL}}
{\footnotesize \footnotemark[4] \cite{WVGT}}
\end{table}

d

Google's MediaPipe for landmark detection \cite{mediapipe} was used to extract coordinate-based input features, enabling a simplified Multi-Layer Perceptron and Gated Recurrent Unit (MLP-GRU) architecture for the SLR models. This approach reduces data requirements and enhances robustness against signer body size variations compared to using raw frames \cite{duy2021vietnamese}. The paper is structured as follows: An introduction to the research contribution is presented in Section 1 whilst salient research from the existing literature is covered in Section 2. Section 3 details the methodology, including data pre-processing, MediaPipe key-points extraction, architecture, as well as TL results. Section 4 presents the ablation study and finally, Section 5 concludes the study by discussing the key insights and findings.

\section{Related Work}
\label{sec: Related Work}
\begin{figure}[htbp!]
    \centering
    \includegraphics[width=0.75\textwidth]{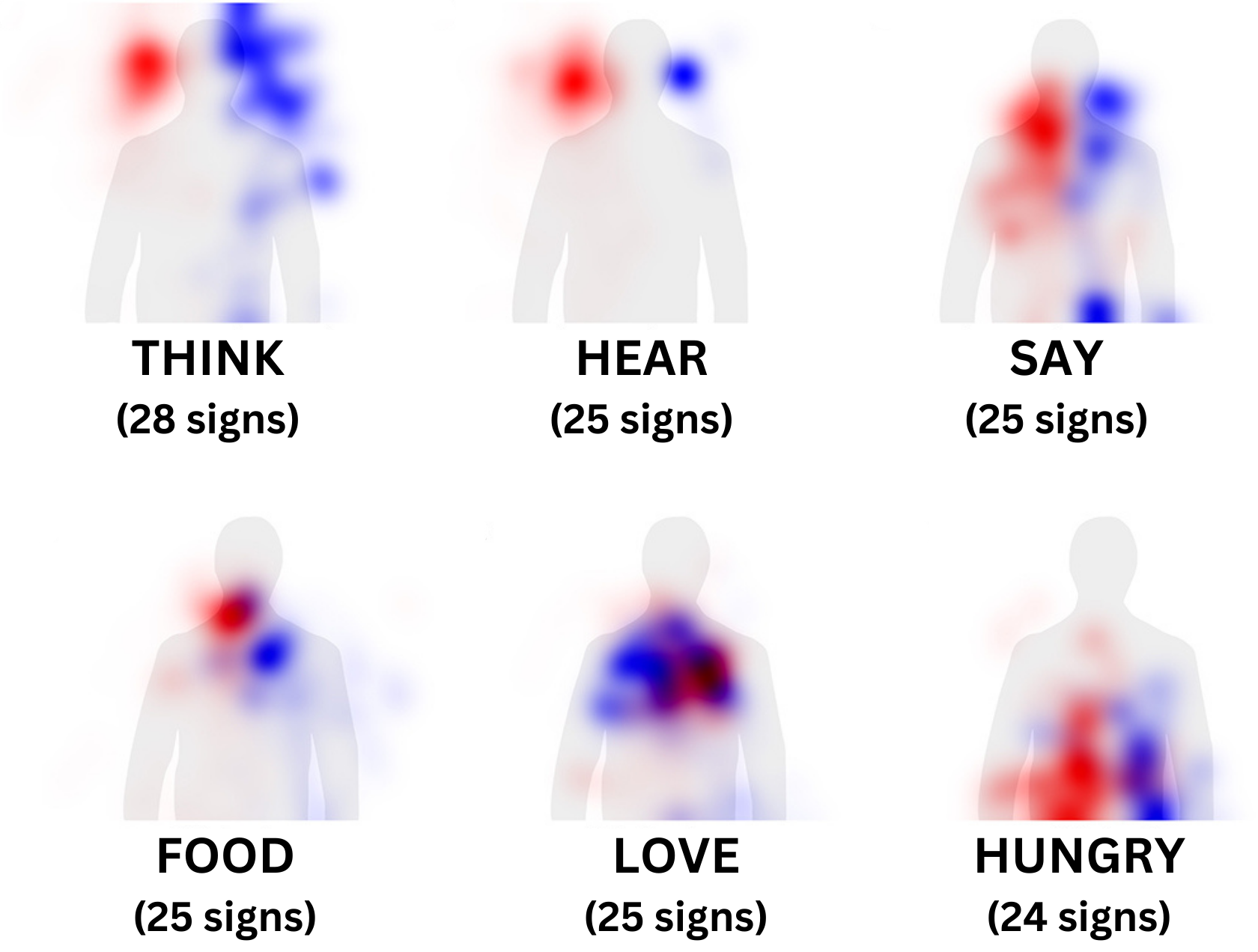}

    \caption{Sign gesture hand activity for individual concepts across languages. {\cite{visualiconicity}}}

    \label{fig:ostling_individualconcept}

\end{figure}

\begin{figure}[htbp!]
    \centering
    \includegraphics[width=0.75\textwidth]{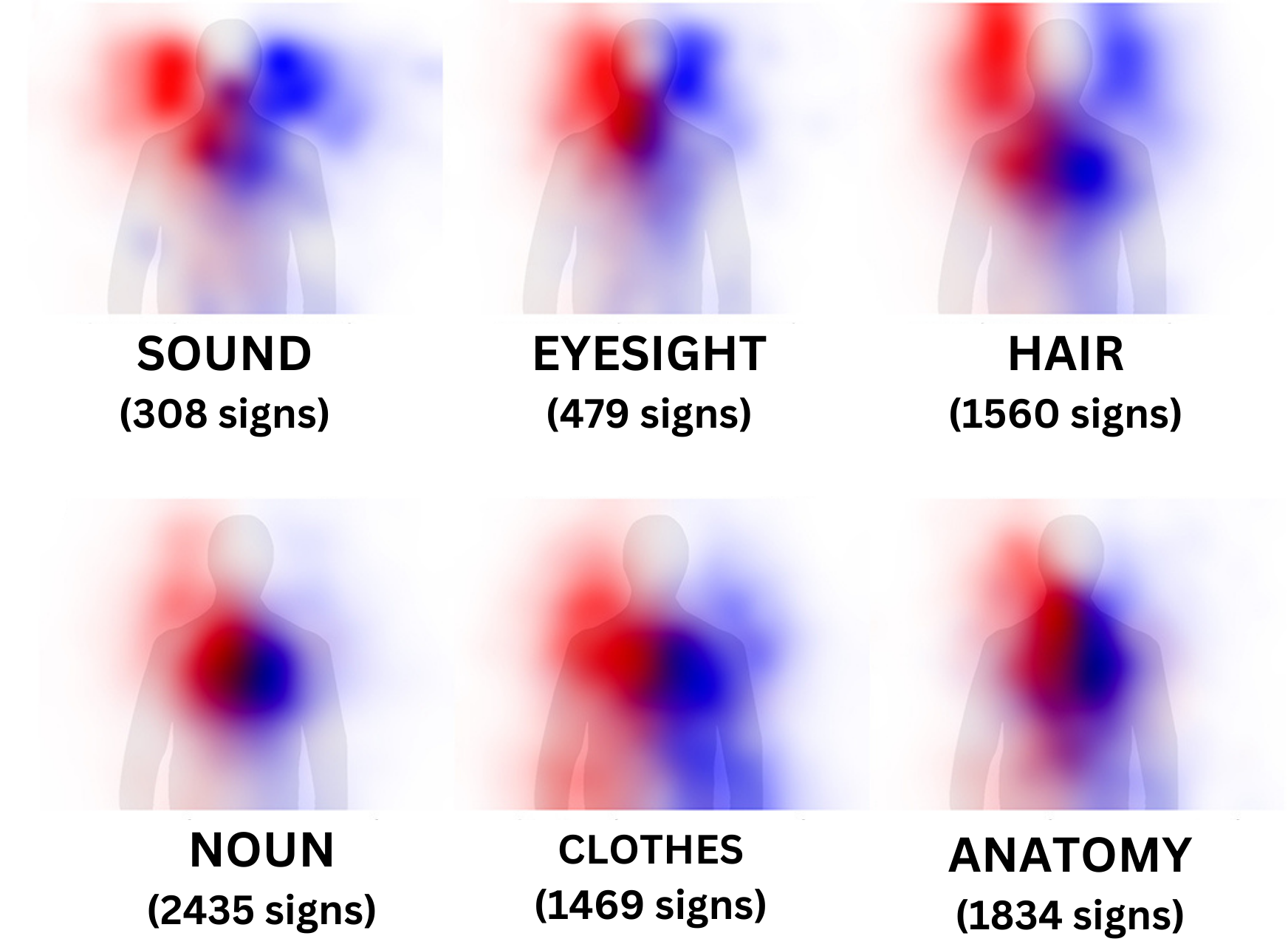}
    \caption{Sign gesture hand activity for groups of concepts across languages. {\cite{visualiconicity}}}

    \label{fig:ostling_largeconcept}

\end{figure}

Recent advancements in non-verbal communication research highlight the effectiveness of TL in enhancing Sign Language Recognition (SLR), especially using pre-trained ImageNet models. Several recent studies \cite{DBLP:journals/corr/abs-1805-06618},\cite{cayamcela_lim_2019}, \cite{https://doi.org/10.48550/arxiv.2010.07827}, \cite{nishat_shopon_2020}, \cite{Zakariah2022SignLR}, \cite{Shania2022TranslatorOI}, \cite{Thakar2022SignLT}, \cite{Das2022AHA}, \cite{Jiang2020FingerspellingIF}, \cite{SHARMA2023119772} endorse this TL approach. Meanwhile, there has been a growing interest in domain-specific TL, as evidenced by studies that transfer knowledge between large corpus languages, such as BSL to the more moderate ASL dataset \cite{Bird2020BritishSL}, or from ASL to AUTSL or LSE to AUTSL \cite{vazquez-etal-2022-latest}, and recently from CSL to ASL, as well as from Argentine Sign Language (LSA) to ASL \cite{Artiaga2023}. However, the studies by \cite{Bird2020BritishSL} and \cite{Artiaga2023} focused on signs present in both their source and target datasets which are often iconic. Conversely, \cite{vazquez-etal-2022-latest} attributed improved recognition accuracy to the similar data acquisition method used by their source and target datasets through Kinect. This research aims to address the gap presented by the lack of studies comparing TL between signs with similarities such as in the case of iconic signs and between signs with no obvious similarities. The selection of signs for this study aligns with the findings of \cite{visualiconicity}, who used automated methods to visualise hand activity patterns across 31 sign languages. These patterns, illustrated in Figures \ref{fig:ostling_individualconcept} and \ref{fig:ostling_largeconcept} reveal similarities in hand movements, particularly for individual concepts. 

While cross-linguistic patterns also emerged for broader lexical or semantic categories, they are less pronounced than those for individual concepts.

\section{ Method}
\label{sec:Technical Approach}
This section provides an extensive account of the data pre-processing, mediapipe keypoint extraction and the MLP-GRU architecture, along with its critical hyper-parameters. It concludes with a comprehensive explanation of the experimental TL results.

\subsection{Data Pre-processing and Extracting MediaPipe Keypoints}
\label{subsec:Data Pre-processing}

Two pairs of isolated sign language datasets are used for this study, namely CSL SLR500 and KArSL, and GSL isol and Woordenboek VGT. Table \ref{tab: Overview of the iconic subsets} presents the class counts for each iconic subset, along with the sample count per class, as well as the iconic concepts these classes belong to. For the training and testing, each subset is randomly split 80\% for training and 20\% for testing with the exception of KArSL as its authors indicated a split for this dataset - approximately 78\% for training and approximately 22\% for testing. For the task of iconic TL, signs were chosen from both datasets based on iconic individual concepts c.f. Fig. \ref{fig:ostling_individualconcept} and larger categories c.f. Fig. \ref{fig:ostling_largeconcept} identified by \cite{visualiconicity}. MediaPipe Holistic Landmarker task (MediaPipe 0.10.3) was used to extract hands, shoulders and wrist landmarks from the source and target dataset video samples. Facial keypoints were excluded since the focus of this study is on word-level SLR, where grammatical and syntactic markers are less relevant, and the signers in both datasets only mouth the equivalent word for the sign. Additionally, facial expressions fall outside the scope of this paper's reference study \cite{visualiconicity} on cross-linguistic iconicity. Each MediaPipe landmark comprised x, y and z coordinates. The x and y coordinates are normalised to the frame dimensions, ranging from 0.0 to 1.0, while the z coordinate indicates depth and is approximately the same scale as x. The coordinates are closer to 1.0 when the hands are at rest. The KArSL, LSFB and GSL isol datasets are pre-processed, with frames lacking activity removed. However, this is not the case with the CSL SLR500 and WoordenboekVGT datasets. Hence, to eliminate outliers from these datasets, keypoints are only extracted from frames where the y-coordinate value of either the left or right-hand wrist was below 0.6.

\begin{figure}[htbp!]

    \centering

    \includegraphics[width=.75\textwidth]{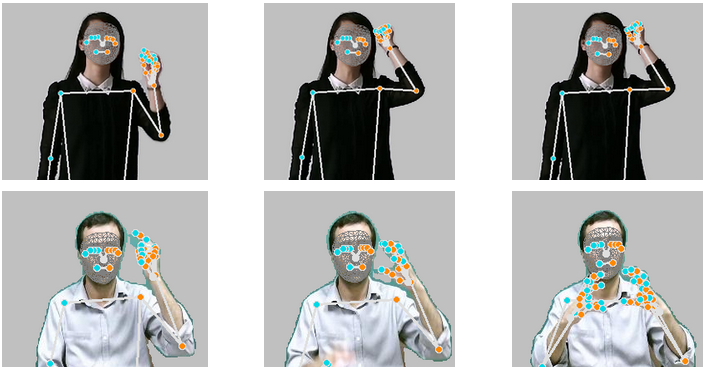}

    \caption{MediaPipe keypoint landmarks detailing the anatomy sign for \emph{Skull} (KArSL) and \emph{Head} (CSL)}

    \label{fig:skull_head}

\end{figure}

Figure \ref{fig:skull_head} shows the MediaPipe keypoint landmarks on select video frames representing the KArSL sign for \emph{Skull} and the CSL sign for \emph{Head}. The cross-linguistic similarities between these two iconic signs are apparent, with the only noticeable difference being the inclusion of right-hand activity towards the end of the KArSL gesture.

\subsection{MLP-GRU}
\label{sec:MLP-GRU}
Both the baseline (non-TL KArSL and non-TL Woordenboek VGT) and transfer learning models (SLR500 to KArSL, and GSL isolated to Woordenboek VGT) in this study shared an MLP-GRU architecture. The MLP learned spatial information for each set of keypoints in a sample, while the GRU extracted temporal information from the features produced by the MLP. The MLP had a single hidden layer, and the GRU consisted of only one recurrent layer. The Rectified Linear Unit (ReLU) activation function, as defined in \ref{eq:relu}, was applied to the MLP layer.
\begin{equation}
ReLU(x) = max(0,x)
\label{eq:relu}
\end{equation}

where \(x \in \mathbb{R} \). Softmax function, as described in \ref{eq:softmax}, is applied to the output layer.

\begin{equation}
\sigma(y_{i}) = \left(\frac{e^{y_{i}}}{ \sum\limits_{j} e^{y_{j}}}\right)
\label{eq:softmax}
\end{equation}

where \(\sigma\) is the softmax, \(y\) is the input vector, \(e^{y_{i}}\) denotes the standard input vector exponential function, \(j\) represents the number of classes, and \(e^{y_{j}}\) is the standard output vector exponential function. Grid search was performed to determine the best number of MLP neurons and GRU hidden size for the baseline target models. The resulting best set of hyper-parameters was then used to build the source models. Table \ref{tab:baseline results} displays the performance of different sets of MLP-GRU hyper-parameters in the baseline KArSL and Woordenboek VGT recognition tasks. As Woordenboek VGT is an imbalanced dataset, the macro F1 score, as shown in \ref{eq:f1score}, is employed in addition to the accuracy metric, as shown in \ref{eq:accuracy}, due to its ability to provide a balanced evaluation of performance across all classes. The formulation for the macro F1 Score  is shown as follows:

\begin{equation}
\small
\text{F1}_{\text{macro}} = \frac{1}{N} \sum_{i=1}^{N} \frac{2 \cdot \text{Precision}_i \cdot \text{Recall}_i}{\text{Precision}_i + \text{Recall}_i}
\label{eq:f1score}
\end{equation}
where \(N\) is the number of classes, \(\text{Precision}_i\) is the precision for class \(i\), and \(\text{Recall}_i\) is the recall for class \(i\).

Meanwhile, the formulation for accuracy is shown below:

\begin{equation}{}
\small
    Accuracy = \frac{TP + TN}{TP + TN + FP + FN}
    \label{eq:accuracy}
\end{equation}
where \(TP\), \(TN\), \(FP\) and \(FN\) represent the true positive, true negative, false positive and false negative values respectively.

It is worth noting that for Woordenboek VGT, two combinations of MLP Neurons - GRU Hidden Size resulted in the same best macro F1 score; however, the 2048-4096 combination was selected as the best set of hyper-parameters as it achieved the best macro F1 score at 2000 epoch, whereas the 2000-3000 combination achieved it at 2362 epochs. 

\begin{table}[!htbp]
\centering

\resizebox{1\textwidth}{!}{%
\begin{tabular}{lllll}
\hline
\textbf{\begin{tabular}[c]{@{}l@{}}Target Subsets \,\,\, \end{tabular}}                    & \textbf{\begin{tabular}[c]{@{}l@{}}MLP Neurons\,\,\, \end{tabular}} & \textbf{\begin{tabular}[c]{@{}l@{}}GRU Hidden Size\,\,\, \end{tabular}} & \textbf{\begin{tabular}[c]{@{}l@{}}Recognition Acc. (\%)\end{tabular}} & \textbf{\begin{tabular}[c]{@{}l@{}}Macro F1 Score (\%)\,\,\, \end{tabular}} \\ \hline
\multirow{5}{*}{KArSL}                                                      & 256                                                   & 512                                                       & 77.81                                                           & -                                                                 \\
                                                                            & 512                                                   & 1024                                                      & 79.53                                                           & -                                                                 \\
                                                                            & 1024                                                  & 2048                                                      & 79.69                                                           & -                                                                 \\
                                                                            & 2000                                                  & 3000                                                      & 80.15                                                           & -                                                                 \\
                                                                            & 2048                                                  & 4096                                                      & 79.22                                                           & -                                                                 \\ \hline
\multirow{5}{*}{\begin{tabular}[c]{@{}l@{}}Woordenboek \\ VGT\end{tabular}} & 256                                                   & 512                                                       & 86.12                                                           & 58.75                                                            \\
                                                                            & 512                                                   & 1024                                                      & 90.28                                                           & 84.35                                                            \\
                                                                            & 1024                                                  & 2048                                                      & 90.21                                                           & 85.92                                                            \\
                                                                            & 2000                                                  & 3000                                                      & 90.28                                                           & 87.88                                                            \\
                                                                            & 2048                                                  & 4096                                                      & 90.28                                                           & 87.88                                                            \\ \cline{1-5} 
\end{tabular}
}
\caption{Impact of MLP neurons and GRU hidden sizes on the accuracy and macro F1 score of the target tasks.}
\label{tab:baseline results}
\end{table}

Adam \cite{DBLP:journals/corr/KingmaB14} optimizer was used to train the models using a batch size of 32 with a learning rate of 1e-05. The training continued for an uncapped number of epochs and was terminated when no improvement in loss occurred for 200 epochs. The loss function used is categorical cross-entropy shown in \ref{eq:crossentropy}. 

\begin{equation}
\small
    Loss = - \frac{1}{N}\sum_{i=1}^N\sum_{j=1}^Jy_{i,j}\log(\hat{y}_{i,j})
    \label{eq:crossentropy}
\end{equation}

where \(N\) is the size of the test set, and \(J\) is the classification category. Utilising an Intel i9-9940X processor, the models were trained on an 11GB NVIDIA GeForce RTX 2080Ti GPU with 128GB of RAM.

\subsection{Transfer Learning (TL) and Results}

TL involves transferring knowledge from one domain to another. In this study, TL was conducted by saving the learned weights from source SLR tasks (iconic CSL and non-iconic CSL recognition) and using them as initial weights for the target task (ArSL recognition), known as weight initialisation - c.f. Figure \ref{fig:architecture}. These weights were then fine-tuned during training until the loss converged to a minimum value. Weight initialisation was applied only to the MLP layer of the architecture since the GRU remained fixed with a single layer. Table \ref{tab:TL_results} displays the recognition accuracy of the fine-tuned KArSL and Woordenboek VGT models, showing improvements over their non-TL baselines.

\begin{figure}[]

    \centering

    \includegraphics[width=0.65\textwidth]{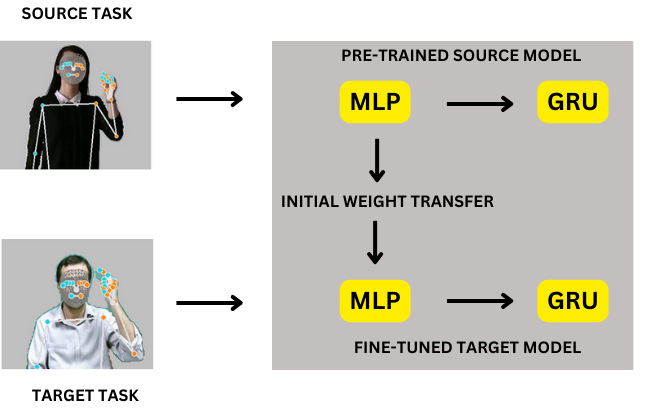}

    \caption{MLP-GRU execution on SLR TL tasks.}

    \label{fig:architecture}

\end{figure}

\begin{table}[]
\normalsize
\centering
\resizebox{.85\textwidth}{!}{%
\begin{tabular}{lll}
\hline
\textbf{Task}                                                         & \textbf{\begin{tabular}[c]{@{}l@{}}Recognition Accuracy (\%)\end{tabular}} & \textbf{\begin{tabular}[c]{@{}l@{}}Macro F1 Score (\%)\end{tabular}} \\ \hline
\begin{tabular}[c]{@{}l@{}}KArSL Baseline\end{tabular}             & 80.15                                                                    & - \\ 
\begin{tabular}[c]{@{}l@{}}CSL SLR500 to KArSL TL\end{tabular}      &      85.78                                                                    & - \\ \hline
\begin{tabular}[c]{@{}l@{}}Woordenboek VGT Baseline\end{tabular}    & 90.28                                                                    & 87.88 \\ 
\begin{tabular}[c]{@{}l@{}}GSL isolated to Woordenboek VGT \,\,\, \end{tabular} &    91.25                                                                       & 87.88 \\ \hline
\end{tabular}
}
\caption{Summary of iconic TL tasks results}
\label{tab:TL_results}
\end{table}

The results indicate that transferring knowledge from iconic CSL SLR500 to KArSL led to a notable 7.02\% increase in recognition accuracy, showcasing the effectiveness of leveraging iconicity for improved performance. Similarly, the transfer learning from iconic GSL isolated to Woordenboek VGT resulted in a modest 1.07\% improvement in accuracy. However, the corresponding macro F1 scores remain unchanged at 87.88\%, indicating that transfer learning did not affect the balance of performance across different classes. Nevertheless, transfer learning helped achieve the same macro F1 score at an earlier epoch of 1867 compared to 2000 with the baseline.

\section{Ablation Study}

\subsection{Effect of pre-training with non-iconic signs}
\label{subsec: Effect of pre-training with non-iconic signs}
A comparison was conducted to assess the performance of using non-iconic signs from the source languages, aiming to determine whether there are differences compared to simply pre-training with non-iconic signs. The pre-processing and experimental setup for the non-iconic transfer were similar to that of the iconic transfer \ref{sec:MLP-GRU}, with an equal number of non-iconic signs selected from each source dataset to ensure a fair comparison.

Table \ref{tab:nonTL_results} presents the results of pre-training with non-iconic signs. It is observed that for CSL SLR 500 to KArSL non-iconic transfer, there was a positive transfer, showing an improvement of 6.84\% from the baseline. This improvement, while lower than the 7.02\% improvement from the iconic transfer, still indicates the effectiveness of this approach. On the other hand, for GSL isolated to Woordenboek VGT, the best accuracy remained similar to that of the baseline; however, the model achieved this accuracy at an earlier epoch of 238 compared to the baseline, which reached this accuracy at 502 epochs. Similarly, the F1 score remained unchanged compared to that of the baseline; however, the model achieved this score at 1815 epochs compared to the baseline, which achieved this score at 2000 epochs. This suggests that for this specific transfer scenario, utilising non-iconic signs led to faster convergence to a similar level of performance compared to the baseline model.

\begin{table}[]
\normalsize
\centering
\resizebox{.85\textwidth}{!}{%
\begin{tabular}{lll}
\hline
\textbf{Task}                                                         & \textbf{\begin{tabular}[c]{@{}l@{}}Recognition Accuracy (\%)\end{tabular}} & \textbf{\begin{tabular}[c]{@{}l@{}}Macro F1 Score (\%)\end{tabular}} \\ \hline
\begin{tabular}[c]{@{}l@{}}KArSL Baseline\end{tabular}             & 80.15                                                                    & - \\ 
\begin{tabular}[c]{@{}l@{}}CSL SLR500 to KArSL TL\end{tabular}      & 85.63                                                                    & - \\ \hline
\begin{tabular}[c]{@{}l@{}}Woordenboek VGT Baseline\end{tabular}    & 90.28                                                                    & 87.88 \\ 
\begin{tabular}[c]{@{}l@{}}GSL isolated to Woordenboek VGT \,\,\, \end{tabular} & 90.28                                                                    & 87.88 \\ \hline
\end{tabular}
}
\caption{Summary of non-iconic TL results}
\label{tab:nonTL_results}
\end{table}

\subsection{Effect of pre-training with iconic+non-iconic signs}
\label{subsec: Effect of pre-training with iconic+non-iconic signs}
Experiments were conducted to assess whether there would be an increase in accuracy for the target tasks when combining iconic signs and non-iconic signs as source tasks. The pre-processing and experimental setup for this ablation study were kept similar to the iconic transfer \ref{sec:MLP-GRU} to ensure a fair comparison. Results obtained from transferring with iconic+non-iconic signs showed some improvement in accuracy for the TL task of CSL SLR500 to KArSL, as presented in Table \ref{tab:iconic+noniconic_results}. However, for the GSL isolated to Woordenboek VGT TL task, the best accuracy remained similar to the baseline, albeit at the epoch of 164, which is earlier than both the baseline and the equivalent non-iconic tasks \ref{subsec: Effect of pre-training with non-iconic signs}, which is at the 595 and 164 epochs, respectively. The macro F1 scores for the Woordenboek VGT tasks, which are consistently at 87.88\%, indicate that while the overall accuracy showed some improvements with transfer learning, the balance and consistency of the model's performance across different classes remained unchanged. However, transfer learning helped achieve the same macro F1 score at an earlier epoch of 1679 compared to 2000 with the baseline (no transfer learning).

\begin{table}[]
\normalsize
\centering
\resizebox{.85\textwidth}{!}{%
\begin{tabular}{lll}
\hline
\textbf{Task}                                                         & \textbf{\begin{tabular}[c]{@{}l@{}}Recognition Accuracy (\%)\end{tabular}} & \textbf{\begin{tabular}[c]{@{}l@{}}Macro F1 Score (\%)\end{tabular}} \\ \hline
\begin{tabular}[c]{@{}l@{}}KArSL Baseline\end{tabular}             & 80.15                                                                    & -- \\ 
\begin{tabular}[c]{@{}l@{}}CSL SLR500 to KArSL TL\end{tabular}      &      84.06                                                                    & -- \\ \hline
\begin{tabular}[c]{@{}l@{}}Woordenboek VGT Baseline\end{tabular}    & 90.28                                                                    & 87.88 \\ 
\begin{tabular}[c]{@{}l@{}}GSL isolated to Woordenboek VGT\end{tabular} &    90.28                                                                       & 87.88 \\ \hline
\end{tabular}
}
\caption{Summary of iconic+non-iconic TL results}
\label{tab:iconic+noniconic_results}
\end{table}

\subsection{Effect of fewer similarities in iconic signs}
An experiment was conducted to compare the performance between language pairs that share fewer than three similar iconic concepts. This experiment focused on the sign language pair Iranian and French-Belgian - as their corresponding SL dataset met the criteria for sharing fewer than three similar iconic concepts. For the source task, the Iranian SL dataset, MedSLset\footnote{\href{https://ieee-dataport.org/open-access/display-multimodal-medslset-medical-sign-language-set}{https://ieee-dataport.org/open-access/display-multimodal-medslset-medical-sign-language-set}}, was selected while for the target task, the French-Belgian dataset Langue des Signes de Belgique Francophone (LSFB-ISOL)\footnote{\href{https://lsfb.info.unamur.be/}{https://lsfb.info.unamur.be/}} was chosen. Table \ref{tab: MedSLset2LSFBdatasetdetails} presents the relevant statistics for this language pair. Although MedSLset has fewer samples per class, it is deemed a better source as it has a balanced number of samples while LSFB-ISOL does not. It is noted that the MedSLset dataset only shares 2 similar iconic concepts with LSFB-ISOL, namely \emph{Anatomy} and \emph{Sound}.

\begin{table}[htbp!]
\centering
\resizebox{1\textwidth}{!}{%
\begin{tabular}{lllll}
\hline
\textbf{Dataset} & \textbf{\begin{tabular}[c]{@{}l@{}}Iconic subset\\ no. of \\classes\end{tabular}} & \textbf{\begin{tabular}[c]{@{}l@{}}Iconic subset\\ no. of \\ samples \\ per class (mean)\end{tabular}} & \textbf{\begin{tabular}[c]{@{}l@{}}Iconic concepts the\\ classes \\ belong to\end{tabular}}                                & \textbf{Balanced} \\ \hline
MedSLset \footnotemark[1] & 23                                                                     & 32                                                                                           & Anatomy, Sound                                                                                                          & Yes               \\ \hline
LSFB    \footnotemark[2] & 42                                                                     & 72.78                                                                                        & \begin{tabular}[c]{@{}l@{}}Anatomy, Sound, \\ Clothes, Food,\\ Think, Eyesight,\\ Say, Love, \\ Hear, Hair\end{tabular} & No                \\ \hline
\end{tabular}
}
\caption{Relevant dataset details for MedSLset and LSFB}
\label{tab: MedSLset2LSFBdatasetdetails}
\vspace{-0.3cm}
{\footnotesize \footnotemark[1]\cite{MedSLset}}
{\footnotesize \footnotemark[2] \cite{LSFB}}
\end{table}

\medskip
\medskip

The pre-processing \ref{subsec:Data Pre-processing} and experimental setup for the non-iconic transfer are similar to the iconic transfer \ref{sec:MLP-GRU} to ensure a fair comparison. Specifically, MedSLset is split to 80\% for training and 20\% for testing while the author-specified split for LSFB was followed. Table \ref{tab:Impact of MLP neurons and GRU hidden sizes} presents the performance of various sets of MLP-GRU hyperparameters in the baseline LSFB model where the configuration with 1024 MLP neurons and 2048 GRU hidden size yielded the best accuracy as well as macro F1 score for the baseline.

\begin{table*}[htbp!]
\centering
\resizebox{.55\textwidth}{!}{%
\begin{tabular}{lllll}
\hline
\textbf{\begin{tabular}[c]{@{}l@{}}Target\\ Subsets\end{tabular}} & \textbf{\begin{tabular}[c]{@{}l@{}}MLP\\ Neurons\end{tabular}} & \textbf{\begin{tabular}[c]{@{}l@{}}GRU Hidden\\ Size\end{tabular}} & \textbf{\begin{tabular}[c]{@{}l@{}}Recognition\\ Acc. (\%)\end{tabular}} & \textbf{\begin{tabular}[c]{@{}l@{}}Macro F1\\ Score (\%)\end{tabular}} \\ \hline
\multirow{5}{*}{LSFB}                                    & 256                                                   & 512                                                       & 58.24                                                           & 11.85                                                        \\
                                                         & 512                                                   & 1024                                                      & 58.66                                                           & 14.00                                                        \\
                                                         & 1024                                                  & 2048                                                      & 58.26                                                           & 11.54                                                        \\
                                                         & 2000                                                  & 3000                                                      & 56.84                                                           & 10.56                                                        \\
                                                         & 2048                                                  & 4096                                                      & 53.76                                                           & 10.05                                                        \\ \cline{1-5} 
\end{tabular}
}
\caption{Impact of MLP neurons and GRU hidden sizes to LSFB}
\label{tab:Impact of MLP neurons and GRU hidden sizes}
\end{table*}

The results of this experiment are displayed in Table \ref{tab: MedSLset2LSFBdatasetdetails} which reveals a case of negative transfer. This outcome confirms the significance of similarity in iconic concepts for successful transfer learning.

\begin{table}[]
\normalsize
\centering
\resizebox{.75\textwidth}{!}{%
\begin{tabular}{lll}
\hline
\textbf{Task}             & \textbf{\begin{tabular}[c]{@{}l@{}}Recognition Accuracy (\%)\end{tabular}} & \textbf{\begin{tabular}[c]{@{}l@{}}Macro F1 Score (\%)\end{tabular}} \\ \hline
LSFB Baseline    & 58.66                                                               & 14.00 \\ 
MedSLset to LSFB \,\,\, & 50.36                                                               & 9.26 \\ \hline
\end{tabular}
}
\caption{Summary of MedSLset to LSFB iconic TL}
\label{tab: MedSLset2LSFBdatasetdetails}
\end{table}

\subsection{Effect of pre-training with ImageNet}

To compare the performance of fine-tuned KArSL and Woordenboek VGT models against the most common approach to TL in SLR, which is ImageNet Pre-training as implemented by \cite{https://doi.org/10.48550/arxiv.2010.07827}, \cite{DBLP:journals/corr/abs-1805-06618}, \cite{cayamcela_lim_2019}, \cite{Das2022AHA}, and \cite{Laines_2023_CVPR}, two sets of experiments were conducted for each target language. These experiments utilised an Imagenet pre-trained convolutional network on top of the GRU. Specifically, we used ResNet50 \cite{7780459} as the pre-trained model for comparison.

\begin{figure}[]

    \centering

    \includegraphics[width=0.65\textwidth]{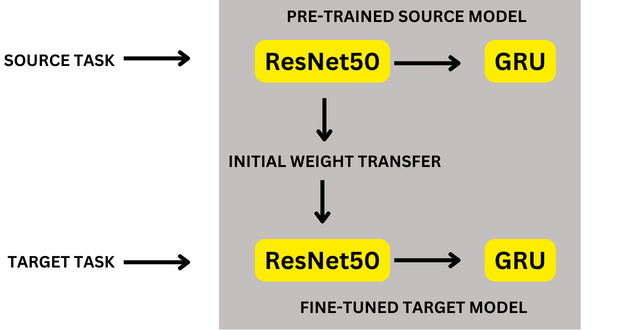}

    \caption{MLP-GRU execution on SLR TL tasks.}

    \label{fig:architecture}

\end{figure}

The results of these experiments are presented in Table \ref{tab:ImageNetPretraining}. It is evident from the results that for the CSL SLR500 to KArSL TL, pre-training with iconic signs yielded better results than pre-training with ImageNet. However, this was not observed for GSL isolated to Woordenboek VGT TL. Nonetheless, these results are not directly comparable because the pre-training with ImageNet uses RGB or raw frames, whereas the iconic TL approach utilises the skeleton keypoints as the input modality.

\begin{table}[]
\centering
\resizebox{.75\textwidth}{!}{%
\begin{tabular}{ll}
\hline
\textbf{Task}                                                                                                 & \textbf{\begin{tabular}[c]{@{}l@{}}Recognition  Acc. (\%)\end{tabular}} \\ \hline
\begin{tabular}[c]{@{}l@{}}iconic CSL SLR 500 to KArSL (keypoints)\end{tabular}                 & 85.78                                                                 \\ \hline
\begin{tabular}[c]{@{}l@{}}ImageNet to KArSL (RGB)\end{tabular}                                & 77.34                                                                 \\ \hline
\begin{tabular}[c]{@{}l@{}}iconic Greek isolated to Woordenboek VGT (keypoints)\end{tabular} & 91.25                                                                 \\ \hline
\begin{tabular}[c]{@{}l@{}}ImageNet to Woordenboek VGT (RGB)\end{tabular}                        & 98.95                                                                 \\ \hline
\end{tabular}
}
\caption{Summary of pre-training with Imagenet in comparison to the iconic TL.}
\label{tab:ImageNetPretraining}
\end{table}

\section{Discussion and Conclusion}

It was demonstrated that transferring knowledge between signs of different languages, regardless of movement similarities, could be highly beneficial, especially where signs shared location boundaries. Specifically, transferring from iconic signs of different source sign languages resulted in the best accuracy for the target sign languages. Specifically, KArSL's recognition accuracy increased from 80.15\% to 85.78\% when initialised with the weights from CSL SLR500 and Woordenboek VGT's recognition accuracy increased from 90.28\% to 91.25\% when initialised with the weights from GSL isolated. Although the macro F1 score remained unchanged for the latter, the use of iconic transfer learning helped achieve this score 133 epochs earlier compared to the baseline. A slight improvement in accuracy is also observed for non-iconic TL and combined iconic and non-iconic TL between CSL SLR500 and KArSL. 

However, from the results of the ablation studies, it can also be concluded that the transferability is sensitive between iconic concepts, such that negative transfer can occur when transferring between datasets containing only a few mutual iconic concepts. Overall, this study provided insight as to how a linguistic concept such as iconicity can help in the automatic recognition of isolated signs which is especially beneficial for low-resource sign languages. Additionally, the resulting pre-trained models from this study can also be instrumental for the more difficult task of sign language translation (SLT). Specifically, the pre-trained target models can be used in training a sign spotter that can recognise individual signs from a video.

\section*{Declarations}

\begin{itemize}
\item \textbf{Funding}:  
This work was conducted with the financial support of the Science Foundation Ireland ADAPT Centre for Digital Content Technology (Grant No.~13/RC/2106). The ADAPT Centre's grant is co-funded under the European Regional Development Fund.

\item \textbf{Conflict of interest / Competing interests}:  
The authors declare that they have no competing interests.

\item \textbf{Code availability}:  
The code used in this study is publicly available at:  
\url{https://github.com/peonycabbage/Iconicity_TransferLearning}

\end{itemize}
\bibliography{iconicity}
\bibliographystyle{splncs04}

\end{document}